\documentclass[letterpaper,10pt]{article}

\usepackage[]{probml}

\ShortHeadings{Latent Diffusion for Missing Data}

\usepackage{blindtext}
\usepackage{wrapfig}
\usepackage{hyperref}
\usepackage{url}

\usepackage{bm,amssymb,amsmath}

\makeatletter
\def\th@plain{%
  \thm@notefont{}
  \itshape 
}
\def\th@definition{%
  \thm@notefont{}
  \normalfont 
}
\makeatother

\def\1{\bm{1}}

\def\rvf{{\mathbf{f}}}

\def\rvm{{\mathbf{m}}}

\def\rvs{{\mathbf{s}}}

\def\rvw{{\mathbf{w}}}
\def\rvx{{\mathbf{x}}}

\def\rvz{{\mathbf{z}}}

\def\mI{{\bm{I}}}

\def\mX{{\bm{X}}}

\DeclareMathAlphabet{\mathsfit}{\encodingdefault}{\sfdefault}{m}{sl}
\SetMathAlphabet{\mathsfit}{bold}{\encodingdefault}{\sfdefault}{bx}{n}

\newcommand{\model}{LDMiss} 

\begin{document}

\title{Latent Diffusion for Missing Data}

\author[1]{Alberte Heering Estad}
\author[1,2]{Ignacio Peis}
\author[1,2]{Jes Frellsen}

\affil[1]{Technical University of Denmark}
\affil[2]{Pioneer Centre for Artificial Intelligence}


\maketitle

\begin{abstract}
	Diffusion models have emerged as powerful generative approaches for missing-data imputation, yet most existing methods operate directly in data space and degrade when training data are heavily incomplete. We investigate whether shifting diffusion to a learned latent representation improves robustness under missing-completely-at-random (MCAR) corruption. To this end, we propose a two-stage framework: a robust VAE-based imputer first learns compact semantic features from incomplete observations, and a diffusion model is then trained in the resulting latent space. Across training missing rates, we perform a controlled comparison against pixel-space diffusion models under the same incomplete-data setting. The latent diffusion model maintains high sample quality and remains stable up to 50\% missingness, while pixel-space diffusion degrades progressively as missingness increases. For downstream imputation, latent diffusion also achieves consistently better performance than pixel-space diffusion. These findings indicate that latent-space modeling mitigates artifact amplification from zero-imputed inputs and provides a more robust generative prior for incomplete-data learning. Overall, our results support latent diffusion as a strong and practically useful alternative to pixel-space diffusion for missing-data problems.
\end{abstract}

\section{Introduction}

Real-world datasets are often incomplete. Missing values arise from from varied sources including loss to follow-up in clinical studies, incomplete documentation in electronic health records, and practical constraints in data collection \citep{wells2013strategies, chang2021prevalence}. The missing data mechanism is typically categorized as missing completely at random (MCAR), missing at random (MAR), or missing not at random (MNAR) \citep{little2002statisticalanalysismissingdata}. Probabilistic generative models offer a principled approach to this problem, as they can learn the data distribution from incomplete observations and impute missing values by sampling from the conditional. Deep generative approaches based on VAEs \citep{nazabal2020handling, peis2022missing,  mattei2019miwaedeepgenerativemodelling}, GANs \citep{yoon2018gainmissingdataimputation, li2018learning} and Flows \citep{richardson2020mcflow} have shown strong results for imputation. More recently, diffusion models have been applied to this task \citep{zhang2025diffputerempoweringdiffusionmodels, givens2025score, 
ouyang2023missdiff, zheng2023diffusionmodelsmissingvalue, tashiro2021csdi}, leveraging the connection between score-based generative modeling and conditional generation \citep{song2021scorebasedgenerativemodelingstochastic}. However, existing diffusion-based methods typically operate in the data space and often assume complete training data. The question of how the choice of either pixel or latent representation space affects both generation quality and imputation performance when training on incomplete data has, to the best of our knowledge, not been studied.

In this work, we present a systematic comparison between diffusion models for missing data in pixel space and our proposed method, denoted \model, which performs diffusion in a latent space learned by a VAE for incomplete data. Both models are trained on MNIST under MCAR missingness at varying rates. We evaluate both generative sample quality and imputation performance. Our results show that \model{} retains higher sample quality as missingness increases and achieves superior imputation performance compared with the pixel-space DDPM.

\section{Score-Based Diffusion Models}
\label{sec:sbdiffusion}

Diffusion models \citep{sohldickstein2015deepunsupervisedlearningusing, song2021scorebasedgenerativemodelingstochastic, ho2020denoisingdiffusionprobabilisticmodels} generate data by progressively corrupting training samples with noise and then learning the reverse denoising process. Denoising Diffusion Probabilistic Models (DDPMs; \citet{ho2020denoisingdiffusionprobabilisticmodels}) formulate this process as a discrete-time Markov chain. In contrast, score-based diffusion models \citep{song2021scorebasedgenerativemodelingstochastic} define a continuous-time forward process $\{\mathbf{x}_{t}\}^{T}_{t=0}$ with $t \in [0,T]$ through the stochastic differential equation (SDE) $\text{d}\mathbf{x}=\mathbf{f}(\mathbf{x},t)\,\text{d}t+g(t)\,\text{d}\mathbf{w}$. The process starts from data $\rvx_0\sim p_0$ and evolves toward a prior distribution $\rvx_T\sim p_T$ that contains no information about $p_0$.

The generative process is characterized by the corresponding reverse SDE: $\text{d}\rvx = \left[ \rvf(\rvx,t)-g(t)^{2}\nabla_{\rvx}\log p_{t}(\rvx) \right] \text{d}t+g(t)\text{d}\rvw
\label{eq:reversesde}$. This reverse SDE depends only on the time-dependent gradient field $\nabla_{\rvx}\log p_{t}(\rvx)$, known as the score, of the transformed data distribution. The score can be estimated by training a time-dependent neural network $\rvs_\theta$ via the Denoising Score-Matching objective \citep{vincent2011connection}:
\begin{align}
    \mathcal{L}(\theta) =
    \mathbb{E}_{t\sim \mathcal{U}(\tau,1)} \left\{ \lambda_{t}\: \mathbb{E}_{\mathbf{x}_{0}\sim p_{0}} \mathbb{E}_{\mathbf{x}_{t}\sim p_{0t}(\mathbf{x}_{t}|\mathbf{x}_{0})} \left[\: \left\| \:\rvs_\theta(\mathbf{x}_{t}, t) - \nabla_{\mathbf{x}_{t}} \log p_{0t}(\mathbf{x}_{t}\: |\: \mathbf{x}_{0}) \:\right\|_2^2 \:\right] \right\}
\end{align}
Once $\rvs_\theta$ is obtained, the reverse SDE can be simulated using numerical methods such as Euler-Maruyama \citep{kloeden1992numericalsolutionsde} to sample new data.

\citet{song2021scorebasedgenerativemodelingstochastic} show that the DDPM perturbation kernels converge to the Variance-Preserving (VP) SDE: $\text{d}\mathbf{x}=-\frac{1}{2}\beta(t)\mathbf{x}\,\text{d}t+\sqrt{ \beta(t) }\,\text{d}\mathbf{w}$, with noise schedule $\beta(t)=\beta_{min}+t(\beta_{max}-\beta_{min})$, when time is continuous. In this study, we use the VP SDE and its corresponding reverse SDE to model the forward and reverse processes, respectively. The linearity of the drift and diffusion coefficients causes the transition kernel $p_{0t}(\rvx_t|\rvx_0)$ to become Gaussian, resulting in a standard Gaussian prior $\rvx_T \sim\mathcal{N}(0,\mI)$. By reparameterization of $\rvx_t\sim \mathcal{N}(\mu,\sigma^2\mathbf I)$, we obtain the simplified objective:
\begin{align}
    \mathcal{L}(\theta)=\mathbb{E}_{t\sim \mathcal{U}(\tau,1),\mathbf{x}_{0}\sim p_{0}, \epsilon\sim\mathcal{N}(0, \mI)} \left[ \lambda_{t} \left\| \mathbf{s}_{\theta}(\mathbf{x}_{t}, t) + \frac{\epsilon}{\sigma_{t}} \right\|^{2}_{2} \right]
    \label{eq:simpleobj}
\end{align}

\section{Latent Diffusion Models}
\label{sec:ldm}

Latent diffusion models (LDMs, \citealp{rombach2022highresolutionimagesynthesislatent}) compress the original images to a latent space of lower dimension, in which the diffusion process is then carried out. This is more computationally efficient and allows the model to train on the most important semantic bits of the data. An LDM consists of an encoder $\mathcal{E}$, which encodes the data $\rvx$ into a latent representation $\rvz = \mathcal{E}(\rvx)$, and a decoder $\mathcal{D}$, that uses the latent to reconstruct the data $\hat\rvx = \mathcal{D}(\rvz) = \mathcal{D}(\mathcal{E}(\rvx))$.

In this study, we use a $\beta$-VAE \citep{higgins2017betavae} with $p(\rvz)=\mathcal{N}(0,\mathbf{I})$ and $q_{\mathcal{E}}(\rvz|\rvx)=\mathcal{N}(\rvz;\mu_{\mathcal{E}},\sigma^{2}_{\mathcal{E}})$, trained with the Evidence Lower Bound (ELBO), with KL regularization weighted by $\beta = 10^{-6}$ to ensure high-fidelity reconstructions. Subsequently, the diffusion model is trained on latent representations $\rvz = \mathcal{E}(\rvx)$ using the objective in Equation \ref{eq:simpleobj}, with $\rvz$ replacing $\rvx$.

\section{Missing Data Framework}

Given dataset $\mX = (\rvx_1,\dots,\rvx_n)^T$ with $p$ features, we define $i \in \{ 1,\dots,n \}$ as the index of the data point $\rvx_i$, and $j \in \{ 1,\dots,p \}$ the index of the feature $\rvx_{ij}$. In the case of MNIST, $\rvx_i$ is a flattened image, and $\rvx_{ij}$ is a pixel in image $\rvx_i$. When working with missing data, we split each data point into missing and observed features $\rvx_i = \{ \rvx^{\text{obs}}_i,\rvx^{\text{miss}}_i \}$. The missing features of each data point $\rvx_i$ are defined according to a binary mask vector $\rvm_i \in \{ 0,1 \}^p$, such that $\rvm_{ij}= 1$, if $\rvx_{ij}$ is observed and $\rvm_{ij}= 0$, if it is missing.
In this study, we assume that the data is \textit{missing-completely-at-random}.

\subsection{Training with Missing Data}

For the diffusion modelling in pixel space, we employ two recent alternatives. First, we adopt the MissDiff approach \citep{ouyang2023missdiff}: after zero-imputing the data, the loss is only computed on observed dimensions by factoring out the missing dimensions and normalizing by the number of observed dimensions:
\begin{align}
    \mathcal{L}(\theta) = \mathbb{E}_{t\sim\mathcal{U}(\tau,1),\rvx_0 \sim p_0,\epsilon\sim\mathcal{N}(0,\mI), \rvm}\left[ \lambda_t\, \frac{|| \rvm \odot(\rvs_\theta(\rvx_t,t) - \nabla_{\rvx_t} \log p_{0t}(\rvx_t|\rvx_0)) ||^2}{\sum_j \rvm_j} \right]
\end{align}
Second, we use DiffPuter \citep{zhang2025diffputerempoweringdiffusionmodels}, an EM-based iterative algorithm that imputes the missing values in the E-step and retrains the diffusion model in the M-step.

For our \model, we follow \cite{mattei2019miwaedeepgenerativemodelling} and zero-impute missing entries. The autoencoder is then trained using a $\beta$-VAE ELBO, where the reconstruction term is evaluated only on the observed data and the KL divergence term is weighted by $\beta_{KL}$. The diffusion process in latent space has the convenient property that missing dimensions factor out under the conditional independence assumption of our diagonal Gaussian encoder (derivation in appendix \ref{app:deriv}). We therefore simply use the latent version of the objective given in Equation \eqref{eq:simpleobj} during latent diffusion.

\section{Self-Guided Missing Data Imputation}
\label{sec:guidance}
\citet{song2021scorebasedgenerativemodelingstochastic} perform imputation by sampling noisy observations from the known forward process $p_t(\rvx^{\text{miss}}_t|\rvx^{\text{obs}}_0)\approx p_t(\rvx^{\text{miss}}_t|\hat{\rvx}^{\text{obs}}_t)$, where $\hat{\rvx}^{\text{obs}}_t$ is a random sample from $p_t(\rvx^{\text{obs}}_t|\rvx^{\text{obs}}_0)$. At each iteration, they replace the observed dimensions with their noisy counterparts and update the missing dimensions using the score network. We refer to this as the \emph{replacement} method. However, this method is not viable for the \model{}, because we have no notion of missing dimensions in latent space.

An alternative for imputing missing values with unconditional diffusion models is to condition the generative process on the observed dimensions. Specifically, we replace the score in Equation \eqref{eq:reversesde} with the conditional score $\nabla_{\mathbf{x}_{t}}\log p_{t}(\mathbf{x}_{t}|\mathbf{x}^{\text{obs}}_0)$. By Bayes' rule, this conditional score decomposes into the unconditional score and a guidance term:
\begin{align}
\label{eq:guidance}
    \nabla_{\mathbf{x}_{t}}\log p_{t}(\mathbf{x}_{t}|\mathbf{x}^{\text{obs}}_0)=\nabla_{\mathbf{x}_{t}}\log p_t(\mathbf{x}_{t})+\nabla_{\mathbf{x}_{t}}\log p_t(\mathbf{x}^{\text{obs}}_0|\mathbf{x}_{t})
\end{align}
However, since $\nabla_{\mathbf{x}_{t}}\log p(\mathbf{x}^{\text{obs}}_0|\mathbf{x}_{t})$ is intractable, we approximate it using Tweedie's formula \citep{efron2011tweediesformula} to obtain the posterior mean estimate $\hat{\mathbf{x}}^{\text{obs}}_{0}(\mathbf{x}_{t})$. Given that our model is Gaussian, the guidance term can be expressed as
\begin{align}
    \log p_t(\rvx^{\text{obs}}_0|\hat{\mathbf{x}}^{\text{obs}}_{0}(\mathbf{x}_{t})) \propto -\frac{1}{2 \sigma^2} \|\mathbf{x}^{\text{obs}}_0-\hat{\mathbf{x}}^{\text{obs}}_{0}(\mathbf{x}_{t})  \|^2,
\end{align}
where we use a standard Gaussian prior $\rvx_T \sim \mathcal{N}(0,\mI)$. For the LDM, we decode the latent posterior mean estimate and compute gradients in pixel-space:
\begin{align}
    \log p_t(\rvx^{\text{obs}}_0|\hat{\mathbf{z}}^{\text{obs}}_{0}(\mathbf{z}_{t})) \propto -\frac{1}{2 \sigma^2} \|\mathbf{x}^{\text{obs}}_0-\mathcal{D}(\hat{\mathbf{z}}^{\text{obs}}_{0}(\mathbf{z}_{t}))  \|^2,
\end{align}
with a standard Gaussian prior $\rvz_T \sim \mathcal{N}(0,\mI)$. We refer to this as the self-guidance method, because the model uses its own predictive capabilities to guide the sampling process. Some approaches use a weight or a schedule on the guidance term \citep{dhariwal2021diffusion}. However, we obtain the best results when scaling the guidance term to have equal magnitude to that of the unconditional score in Equation \ref{eq:guidance}. This ensures equal strength guidance at all time steps.

\section{Experiments}
In this section, we report experiments on the MNIST dataset \citep{deng2012mnist}. For simplicity, we refer to the score-based DDPM described in section \ref{sec:sbdiffusion} as DDPM. The full experimental setup is reported in appendix \ref{app:exsetup}, and the code for reproducing our experiments is accessible at \href{https://github.com/a-estad/latent-diffusion-missing-data}{this location}.

\subsection{Sample Quality}

\begin{figure}[t]
    \centering
    \includegraphics[width=0.75\linewidth]{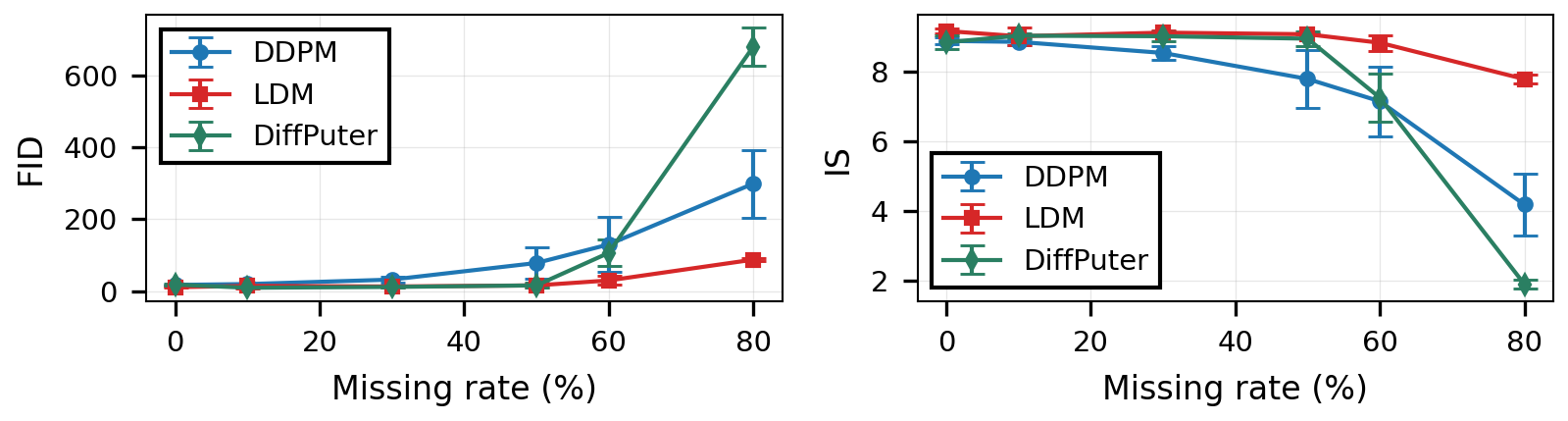}
    \caption{Sample quality (FID, IS) versus training missing rate. Metrics averaged over three seeds.}
    \label{fig:samplequality}
\end{figure}

Figure \ref{fig:samplequality} compares the generative capabilities of each model across training missing rates. The \model{} retains better FID and IS than DDPM across all missing rates, demonstrating greater robustness to incomplete training data. Notably, the sample quality of \model{} and Diffputer remains largely stable up to 50\% missing data, whereas DDPM degrades continuously. In high-missingness regimes, \model{} remains more stable than the alternatives. The \model{} also exhibits lower variance across seeds, suggesting more consistent performance.

\begin{figure}[h!]
    \centering
    \includegraphics[width=\linewidth]{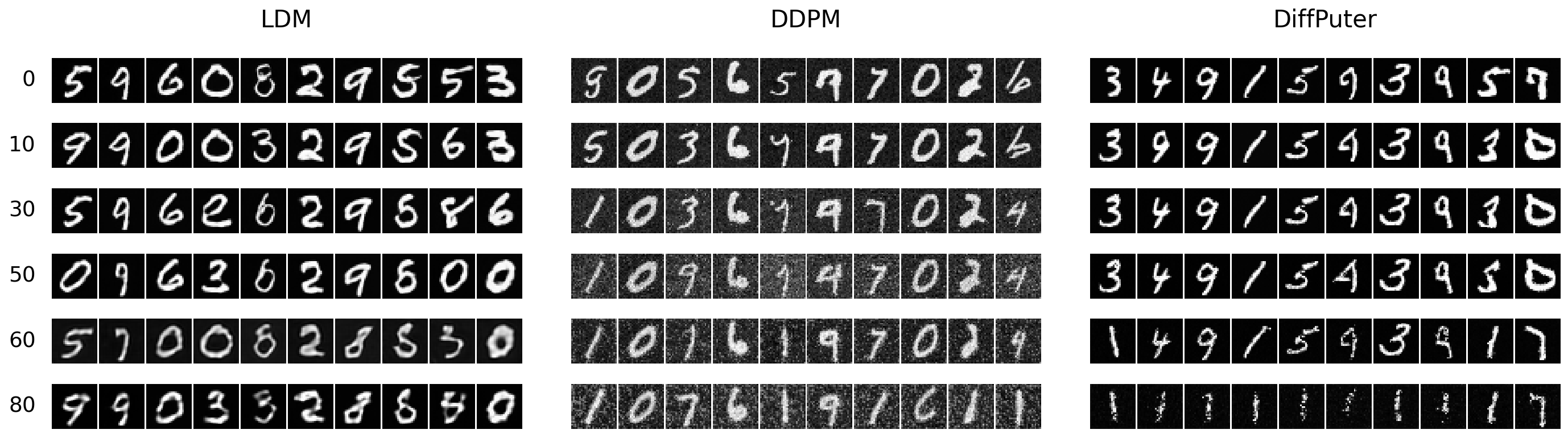}
    \caption{Generated samples across training missing rates.}
    \label{fig:samplegrid}
\end{figure}

Figure \ref{fig:samplegrid} provides a qualitative comparison. The \model{} remains coherent up to 50\% missingness and blurs only at higher rates, while DDPM becomes grainy beyond 10\% because its pixel-space score network directly processes many zero-imputed values. In contrast, \model{} diffuses in compressed latent features, reducing zero-imputation artifacts. DiffPuter is sharp at low to moderate missingness, but in the 60--80\% regime its diversity and fidelity drop.

\subsection{Imputation Quality}
\begin{figure}[t]
    \centering
    \includegraphics[width=1\linewidth]{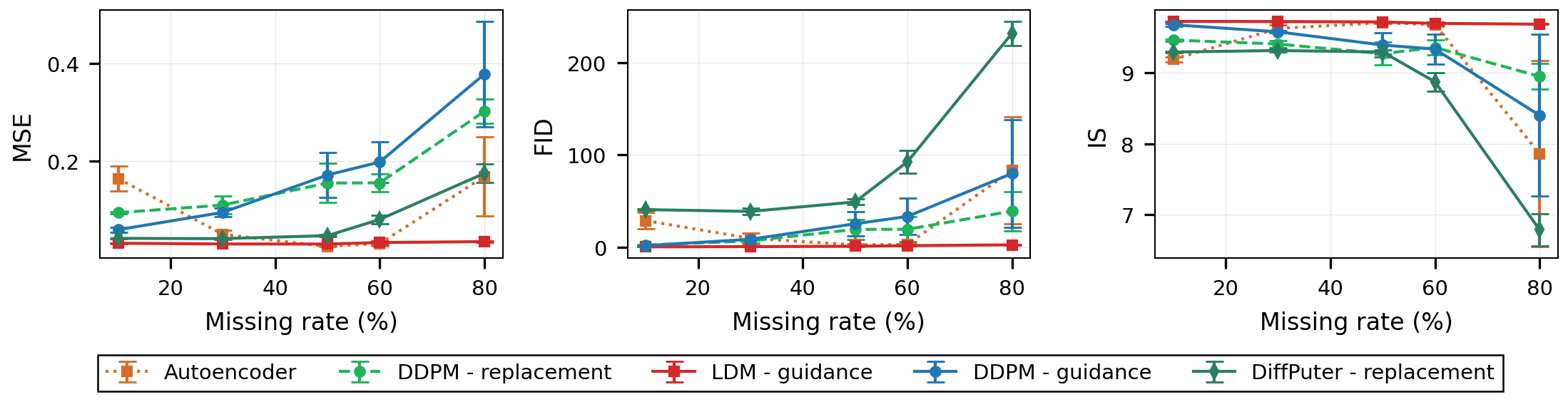}
    \caption{Imputation metrics (MSE, FID, IS) on 10,000 samples at 50\% test missing rate 
versus training missing rate, averaged over three seeds.}
    \label{fig:imputequality}
\end{figure}

Figure \ref{fig:imputequality} compares imputation performance across several methods: self-guidance for \model{} and DDPM, DDPM replacement, and autoencoder reconstruction. All methods are evaluated at a 50\% test missing rate. 
The \model{} remains robust, maintaining strong imputation performance even when trained with 80\% missing data. DDPM degrades as training missingness increases; within DDPM, guidance outperforms replacement at low missing rates but deteriorates faster at high missingness because it relies fully on a degraded score network. DiffPuter, designed for tabular data imputation, is competitive in terms of MSE at low to moderate training missing rates, but at 80\% missingness its reconstructions lose detail and become less consistent than \model{}. Longer EM routines could improve DiffPuter at high missingness, albeit with a substantial increase in computational cost. The autoencoder alone reaches MSE comparable to LDM guidance around 50--60\% training missing rates, but its fixed reconstruction mapping generalizes poorly outside that regime. Overall, \model{} provides the most reliable imputation quality across all  missing rates.

\begin{figure}[h!]
    \centering
    \includegraphics[width=\linewidth]{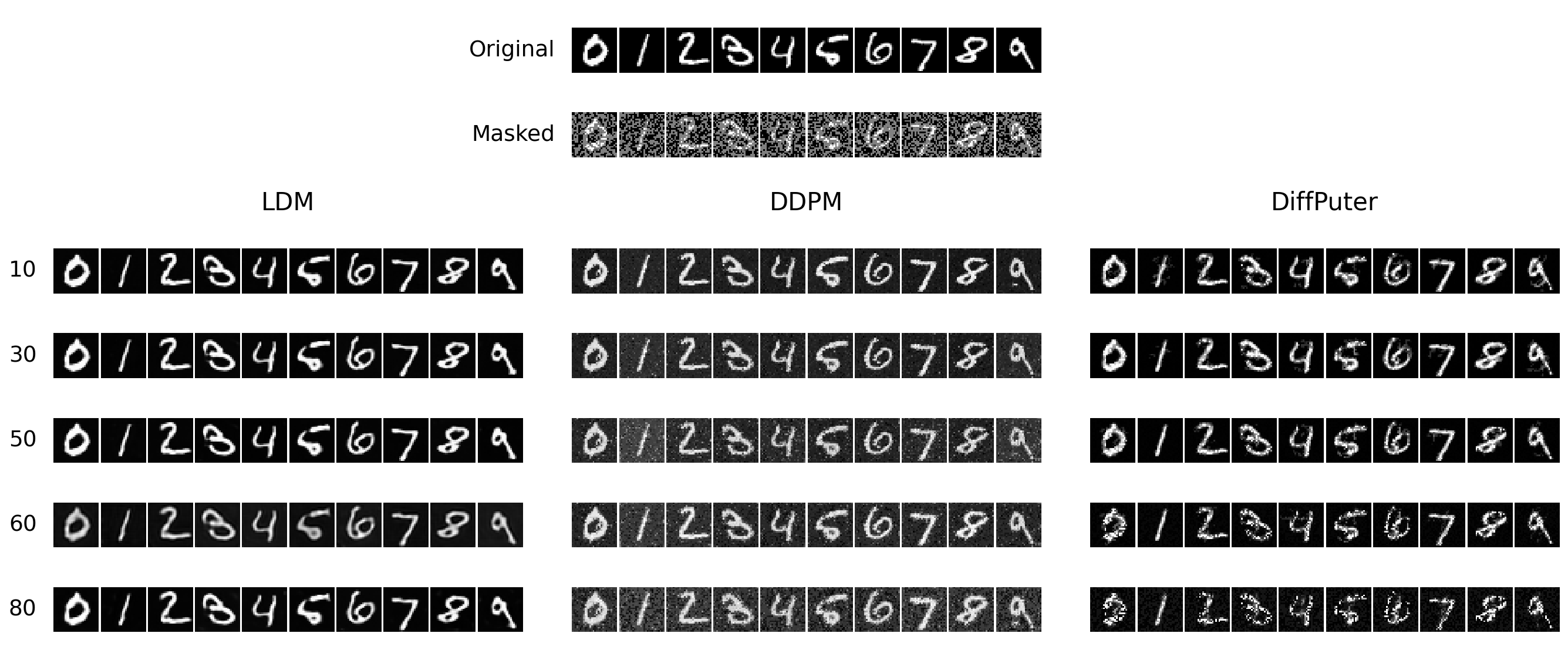}
    \caption{Imputed digits at 50\% test missing rate across training missing rates. 
Left: LDM guidance. Center: DDPM replacement. Right: DiffPuter Replacement.}
    \label{fig:imputationgrid}
\vspace{-0.2cm}
\end{figure}

Figure \ref{fig:imputationgrid} confirms these findings qualitatively. The \model{} imputations remain accurate across all missing rates, while DDPM imputations exhibit the same graininess observed in sample generation.

\section{Conclusion}

We compared pixel-space and latent-space score-based diffusion models trained on incomplete data and found that the \model{} is consistently more robust to training missingness, preserving sample quality up to 50\% missing data and outperforming DDPM on imputation across all training missing rates. A likely reason is that DDPM operates directly in pixel space, where zero-imputed values degrade score estimates, while latent diffusion filters many of these artifacts through semantic compression before denoising. This study is limited to MNIST under MCAR with zero-imputation, and architectural differences may also contribute to the observed gap. Future work should test more complex datasets, MAR/MNAR mechanisms, and alternative imputation functions to assess how well these advantages scale.

\section*{Acknowledgements}

This research was supported by the Villum Foundation
through the Synergy project number 50091, by the Novo
Nordisk Foundation through the Center for Basic Machine Learning Research in Life Science (MLLS, grant no.\@ NNF20OC0062606), and by the Independent Research Fund Denmark (grant no.\@ 5334-00122B and 5334-00076B). Jes Frellsen was further supported by funding from the Reinholdt W. Jorck
og Hustrus Fond. Ignacio Peis acknowledges support by Danish Data Science Academy, which is funded by the Novo Nordisk Foundation (NNF21SA0069429). We thank Pierre-Alexandre Mattei, Federico Bergamin and Hugo Senetaire for valuable and insightful discussions.

\bibliography{references}

\clearpage

\appendix
\section*{Appendix}
\appendix
\section{Experimental Setup}
\label{app:exsetup}
\subsection{Architecture}
\subsubsection{Score Network}
We use a DDPM++ architecture based on \citet{ho2020denoisingdiffusionprobabilisticmodels} and \citet{song2021scorebasedgenerativemodelingstochastic}. The network is a U-Net with sinusoidal timestep embeddings, group normalization, and rescaled skip connections. We trained on MNIST with the configuration reported in table \ref{tab:architectures}.

\begin{table}[h!]
\caption{Model Architectures.}
\label{tab:architectures}
\begin{center}
    \begin{tabular}{rcc}
    \multicolumn{1}{l}{\bf Hyperparameter} & \multicolumn{1}{r}{\bf Pixel-space} & \multicolumn{1}{r}{\bf Latent-space} 
    \\ \hline \\
    Base channels & 48 & 64 \\
    Channel multipliers & 1, 2, 4 & 1, 2 \\
    Residual blocks per resolution & 3 & 2 \\
    Attention resolutions & None & 7 \\
    Dropout & $\sim$ 0.12 & $\sim$ 0.08 \\
    Group normalization groups & 16 & 16 \\
    \end{tabular}
\end{center}
\end{table}

For DiffPuter, we used the architecture specified by \cite{zhang2025diffputerempoweringdiffusionmodels} in their Appendix D.4.

\subsubsection{Autoencoder}
For latent diffusion, we train a KL-regularized VAE following \citet{rombach2022highresolutionimagesynthesislatent}. The encoder and decoder use convolutional residual blocks with channel multipliers [1, 2, 4], downsampling $28 \times 28$ images to $7 \times 7$ latent representations with 2 channels: $1 \times28 \times28 \to 2 \times7 \times7 = 784 \to 98$, resulting in a compression factor of $785/98=8$. We use a very small KL weight ($\beta_{\text{KL}}=10^{-6}$) to ensure high reconstruction quality. The autoencoder is trained with MSE reconstruction loss on data with MCAR missing mechanism.

\subsubsection{Classifier}
\label{app:classifier}
For evaluation, we train a CNN classifier on MNIST with two convolutional layers (32 and 64 channels, $3 \times 3$ kernels) followed by max pooling, and two fully connected layers (64 hidden units). The classifier achieves 99.2\% test accuracy.

\subsection{Training}
\label{app:training}
All models are trained with Adam optimizer using default hyperparameters. We use batch size 256 and train for 50 epochs. Learning rates were determined via hyperparameter sweep to be $\eta_{\text{DDPM}}\approx1.96\cdot10^{-5}$ and $\eta_{\text{LDM}}\approx9.05\cdot10^{-5}$. Each model has been trained for three different seeds: 42, 43, and 44.

\subsubsection{Noise Schedule}
We use the VP SDE with linear $\beta$-schedule:
\begin{align}
    \beta(t)=\beta_{\text{min}}+t(\beta_{\text{max}}-\beta_{\text{min}}), \;\;\;\;\;\;t \in [\tau,1]
\end{align}
with $\beta_{\text{min}} = 0.1$ and $\beta_{\text{max}}=20.0$, following \citet{song2021scorebasedgenerativemodelingstochastic}. We discretize with $T=1000$ timesteps and use $\tau=10^{-3}$ as the minimum time for numerical stability.

\subsubsection{Diffusion in Latent Space}
When traing diffusion models in latents space, we scale the latent space according to \citet{rombach2022highresolutionimagesynthesislatent}.

\subsubsection{Missing Data}
We train models on MNIST with MCAR missing mechanism at rates \{0.0, 0.1, 0.3, 0.5, 0.6, 0.8\}. Missing pixels are set to zero during training (zero imputation).

\subsection{Generation Quality Metrics}
We evaluate sample quality using IS \citep{salimans2016improvedtechniquestraininggans} and FID \citep{heusel2018ganstrainedtimescaleupdate}. IS measures both diversity and clarity of generated images, reflecting how easily they can be classified as distinct digits, with 10 being the highest score. FID measures the similarity between the distributions of generated and real images, capturing how much generated samples resemble the MNIST test set. For imputation, we use FID and IS to assess the quality of imputed images, and additionally MSE between imputed and original images to measure reconstruction accuracy.




\subsection{Compute}
All DDPM experiments have been performed on a NVIDIA Titan V GPU, and all LDM experiments on a NVIDIA GeForce RTX 4070 Ti SUPER GPU.

\section{Missing Dimensions Factorization}
\label{app:deriv}

\begin{align}
    p(\mathbf{x}^{\text{obs}}|\mathbf{z})&=\int p(\mathbf{x}^{\text{obs}},\mathbf{x}^{\text{miss}}|\mathbf{z})\;\text{d}\mathbf{x}^{\text{miss}} \\
    &= \int \prod_{i}p(\mathbf{x}_{i}^{\text{obs}},\mathbf{x}_{i}^{\text{miss}}|\mathbf{z})\;\text{d}\mathbf{x}^{\text{miss}} \\
    &=\prod_{i}p(\mathbf{x}^{\text{obs}}_{i}|\mathbf{z})\cdot \int \prod_{i}p(\mathbf{x}_{i}^{\text{miss}}|\mathbf{z})\;\text{d}\mathbf{x}^{\text{miss}} \label{eq:conditionalindepenceass} \\
    &=\prod_{i}p(\mathbf{x}^{\text{obs}}_{i}|\mathbf{z})\cdot  \prod_{i} \underbrace{\int p(\mathbf{x}_{i}^{\text{miss}}|\mathbf{z})\;\text{d}\mathbf{x}_{i}^{\text{miss}}}_{=\,1} \\
    &= \prod_{i}p(\mathbf{x}^{\text{obs}}_{i}|\mathbf{z}) \\
    &=p(\mathbf{x}^{\text{obs}}|\mathbf{z})
\end{align}
In Equation \eqref{eq:conditionalindepenceass}, we assume conditional independence between observed and missing dimensions given $\rvz$, which holds because we assume a diagonal Gaussian distribution for our encoder $\mathcal{E}$.

\end{document}